\documentclass[10pt,twocolumn,letterpaper]{article}

\usepackage{cvpr}              

\usepackage{graphicx}
\usepackage{amsmath}
\usepackage{amssymb}
\usepackage{booktabs}
\usepackage{color}
\usepackage{enumerate}
\usepackage{float}
\usepackage{xcolor}
\usepackage{times}
\usepackage{epsfig}
\usepackage{multirow}
\usepackage{subcaption}
\usepackage{tabularx}
\usepackage{booktabs}
\usepackage[font=small]{caption}
\usepackage[pagebackref=true,breaklinks=true,colorlinks,bookmarks=false]{hyperref}
\usepackage{float}
\makeatletter
  \newcommand\figcaption{\def\@captype{figure}\caption}
  \newcommand\tabcaption{\def\@captype{table}\caption}
\makeatother

\usepackage[capitalize]{cleveref}
\crefname{section}{Sec.}{Secs.}
\Crefname{section}{Section}{Sections}
\Crefname{table}{Table}{Tables}
\crefname{table}{Tab.}{Tabs.}
\renewcommand{\paragraph}[1]{\vspace{1.25mm}\noindent\textbf{#1}}

\def\cvprPaperID{136} 
\def\confName{CVPR}
\def\confYear{2022}

\begin{document}

\title{GLaMa: Joint Spatial and Frequency Loss for General Image Inpainting}


\author{
    Zeyu Lu$\,^{\dagger}$, Junjun Jiang\thanks{Corresponding author (jiangjunjun@hit.edu.cn).} $\,^{\dagger}$, Junqin Huang$\,^{\ddag}$, Gang Wu$\,^\dagger$, Xianming Liu$\,^\dagger$\\
    $\,^{\dagger}$ Harbin Institute of Technology\quad
    $\,^\ddag$ Beihang University\\
}

\maketitle

\begin{abstract}
The purpose of image inpainting is to recover scratches and damaged areas using context information from remaining parts. In recent years, thanks to the resurgence of convolutional neural networks (CNNs), image inpainting task has made great breakthroughs. However, most of the work consider insufficient types of mask, and their performance will drop dramatically when encountering unseen masks.
 To combat these challenges, we propose a simple yet general method to solve this problem based on the LaMa image inpainting framework \cite{9010689}, dubbed \textbf{GLaMa}. Our proposed GLaMa can better capture different types of missing information by using more types of masks. By incorporating more degraded images in the training phase, we can expect to enhance the robustness of the model with respect to various masks. In order to yield more reasonable results, we further introduce a frequency-based loss in addition to the traditional spatial reconstruction loss and adversarial loss. In particular, we introduce an effective reconstruction loss both in the spatial and frequency domain to reduce the chessboard effect and ripples in the reconstructed image. Extensive experiments demonstrate that our method can boost the performance over the original LaMa method for each type of mask on FFHQ \cite{DBLP:journals/corr/abs-1812-04948}, ImageNet \cite{5206848}, Places2 \cite{zhou2017places} and WikiArt \cite{Saleh2015LargescaleCO} dataset. The proposed GLaMa was ranked first in terms of PSNR, LPIPS \cite{8578166} and SSIM\cite{1284395} in the NTIRE 2022 Image Inpainting Challenge Track 1 Unsupervised \cite{romero2022ntire}.
\end{abstract}

\section{Introduction}
\label{sec:intro}
Image inpainting, also known as image completion, has always been regarded as a challenge to fill the missing area of the image.
Image inpainting can deal with various problems encountered in the real world, such as removing objects in photos, repairing damaged photos \cite{wan2020bringing} or expanding photos.
At the same time, image inpainting needs to maintain the coordination and semantic consistency between the repaired area and remaining parts of the image.
Therefore, image inpainting also calls for strong generation ability. Nowadays, it has become a fundamental research topic in the field of computer vision and image processing society.

\begin{figure}[!t]
    \centering
    \includegraphics[width=0.98\linewidth]{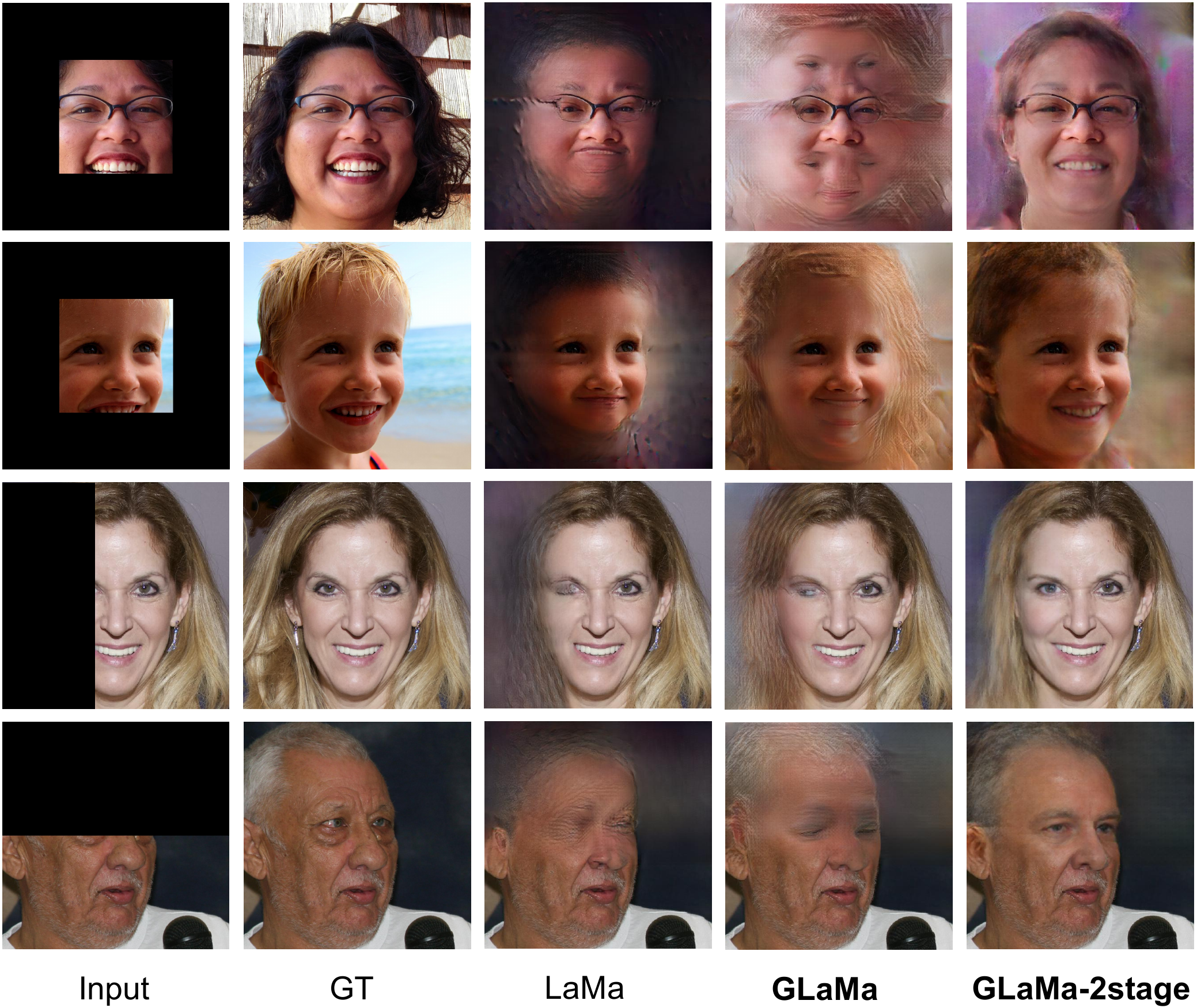}
    \figcaption{Comparison with the LaMa\cite{9707077}, GLaMa and two-stage GLaMa for face Inpainting over some mask settings on FFHQ\cite{DBLP:journals/corr/abs-1812-04948} dataset in $1024\times1024$ resolution.}
    \label{Figure 4}
\end{figure}

Thanks to the fast development of deep learning, the recovered results of these deep learning-based image inpainting approaches are getting better and better.
Recent years, most state-of-the-art approaches are mainly based on convolutional neural networks or transformer.
In the approaches of \cite{9010689, Zeng2022AggregatedCT, Liu2020RethinkingII, zhao2021comodgan}, they apply the convolutional neural networks for image inpainting, while other line of research \cite{wan2021high, yu2021diverse} leverages the transformer in image inpainting at the low-resolution image space, and then introduces the GAN based networks for high quality image generation.
Suvorov et al. \cite{9707077} utilize the Fast Fourier Convolution (FFC) instead of regular convolution to obtain features of global receptive fields in frequency domain.
Most of the current work based on the semantic consistency (with the surrounding areas) can handle the ``background completion'' or ``object removing'' task very well.

Generally, these methods can effectively deal with some common image inpainting tasks.
However, they still face some challenges.
The trained model needs to be able to deal with degraded image with various forms of mask, such as thin strokes, rectangle, and even extreme masks, because we do not know what image degradation process we will encounter.
This is a challenging task because most of methods use specific masks for training, which may acquire poor results when meeting other masks that does not appear in the training processes.
Recently, Andreas et al. \cite{DBLP:journals/corr/abs-2201-09865} address the above issues. Some other diffusion models \cite{Song2021ScoreBasedGM, SohlDickstein2015DeepUL} can also deal with this challenge.
However, it does not mean that the models based on convolutional neural networks (CNNs) or transformer cannot solve this problem well.
In this work, we use more kinds of mask to enhance the robustness of the model.
As shown in Fig. \ref{Figure 3} and \ref{Figure 5}, our method can generate more realistic images.

\begin{figure}[!t]
    \centering
    \includegraphics[width=0.98\linewidth]{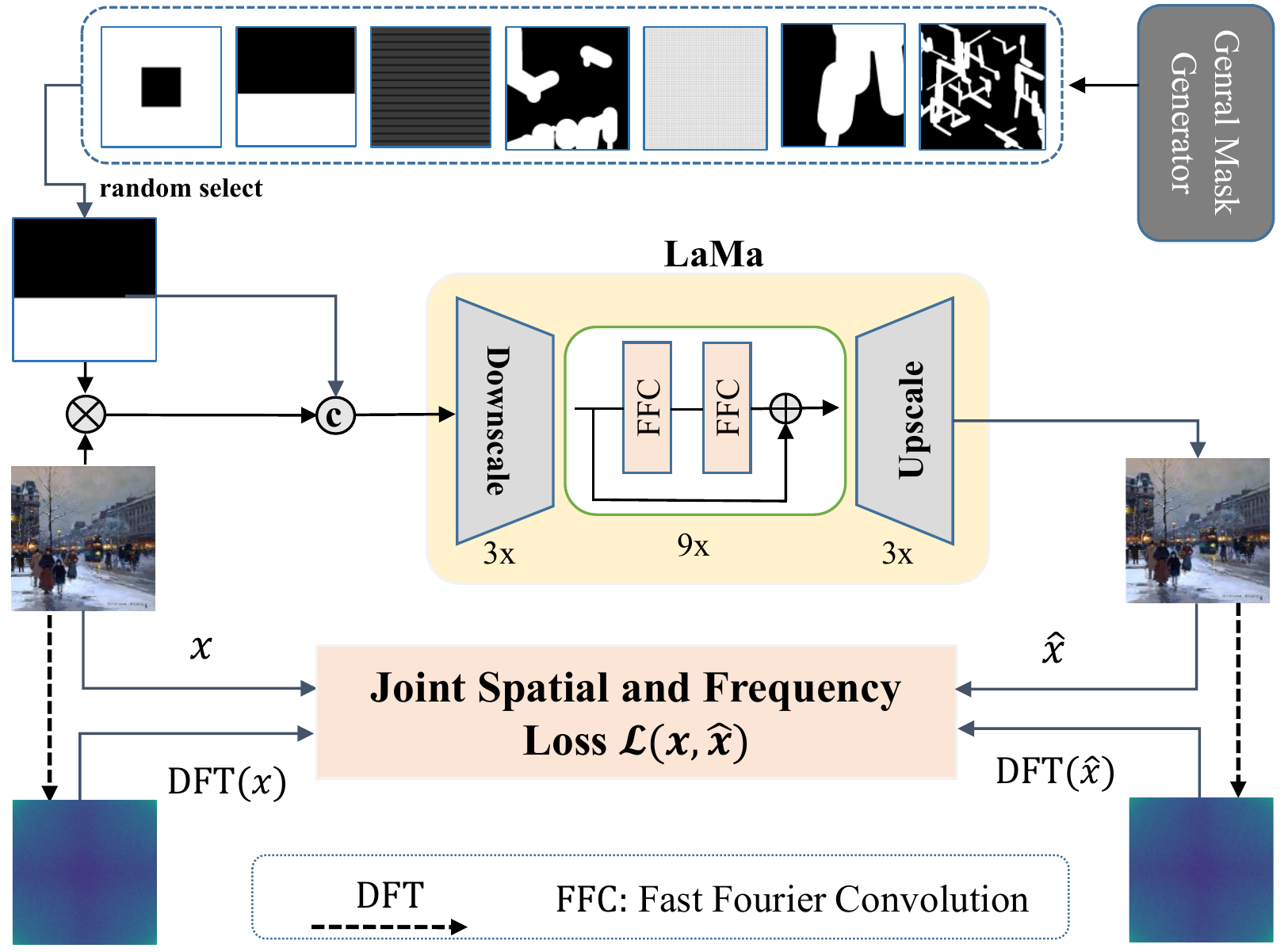}
    \figcaption{Overview of our proposed GLaMa.}
    \label{Figure 2}
\end{figure}

When we visualize the spectra of real and fake images, we find that the difference between spectra of real pictures and fake images is obvious as shown in Fig. \ref{Figure 6}. It can be seen that there are many obvious errors in the spectra of the images generated by LaMa\cite{9707077}.
In addition, by looking closely, we can find that there will be many distinctive checkerboard effects and ripples on the images generated by LaMa \cite{9707077}.
Based on these findings, we add the reconstruction loss in the frequency domain as a regularization term to reduce the checkerboard effect and ripple of the generated image.
We propose a joint spatial and frequency loss to train our network.
As shown in Fig. \ref{Figure 6}, our method can do better in the frequency domain and produce more realistic results at the same time.

To demonstrate our superiority, we compare our method with state-of-the-art image inpainting approach \cite{9707077} on multiple datasets.
As shown in Tabs. \ref{Table 1} and \ref{Table 2}, our method can boost the performance for each type of mask on FFHQ \cite{DBLP:journals/corr/abs-1812-04948}, ImageNet \cite{5206848}, Places2 \cite{zhou2017places} and WikiArt \cite{Saleh2015LargescaleCO} datasets with the same network architecture and training epochs as LaMa \cite{9707077}.
And it was also ranked first on metric PSNR, LPIPS \cite{8578166} and SSIM \cite{1284395} in the NTIRE 2022 Image Inpainting Challenge Track 1 Unsupervised \cite{romero2022ntire}.
The main contributions can be summarized as follows:
\begin{itemize}
    \item We propose to explore the types of mask used in the training process. At the same time, using our mask generation strategy can effectively improve the results of the model.
	\item We use joint spatial and frequency loss in spatial domain and frequency domain with a regular term to reconstruct the image.
	\item As demonstrated in the experiments, our method can achieve significant performance improvements over LaMa \cite{9707077} without changing the model architecture.
\end{itemize}

\section{Related Work}
\label{sec:related}
\subsection{Image Inpainting}
Early work of image inpainting \cite{Criminisi2003ObjectRB, Hays2007SceneCU, Ballester2001FillinginBJ, Bertalmo2003SimultaneousSA, DBLP:conf/siggraph/BertalmioSCB00} are model-driven.
They explore how to fill the missing information by exploiting clues from the local patch or neighbor patches in the input image.

Pathak et al. \cite{Pathak2016ContextEF} propose the first deep learning image inpainting method that utilizes the CNN with an encoder-decoder architecture trained in the same way as GAN \cite{Goodfellow2014GenerativeAN}.
After that, a lot of methods based on CNN have been proposed.
Iizuka et al. \cite{Iizuka2017GloballyAL} improve the performance by exploiting a local-global discriminator, while Yu et al. \cite{Liu2018ImageIF} utilize a contextual attention model to model the long-distance context correlations.
Due to the mask has negative influence on the results when using the regular convolution, several work modify the convolution operator, introducing partial-convolution \cite{Yu2018GenerativeII} conv, gated-convolution \cite{9010689}.
\cite{Zhu2021ImageIB} propose a mask awareness method using cascaded refinement network.
Zeng et al. utilize ``AOT Block'' and ``SoftGAN'' to enhance the generator and discriminator  \cite{Zeng2022AggregatedCT}.
A new GAN called ``Co-Modulated GAN'' combining conditional GAN and modulated GAN is introduced in \cite{Zhao2021LargeSI}.

There are also some work focusing on the fusion of local and global information.
\cite{Liu2020RethinkingII} utilizes feature equalization to fuse local and global features.
Suvorov et al. \cite{9707077} introduce the Fast Fourier Convolution (FFC) \cite{Chi2020FastFC} and regular convolution to obtain features of global and local receptive fields in the frequency domain.
These methods based on CNN could generate reasonable contents for masked regions.

\begin{figure*}[!t]
    \centering
    \includegraphics[width=0.98\linewidth]{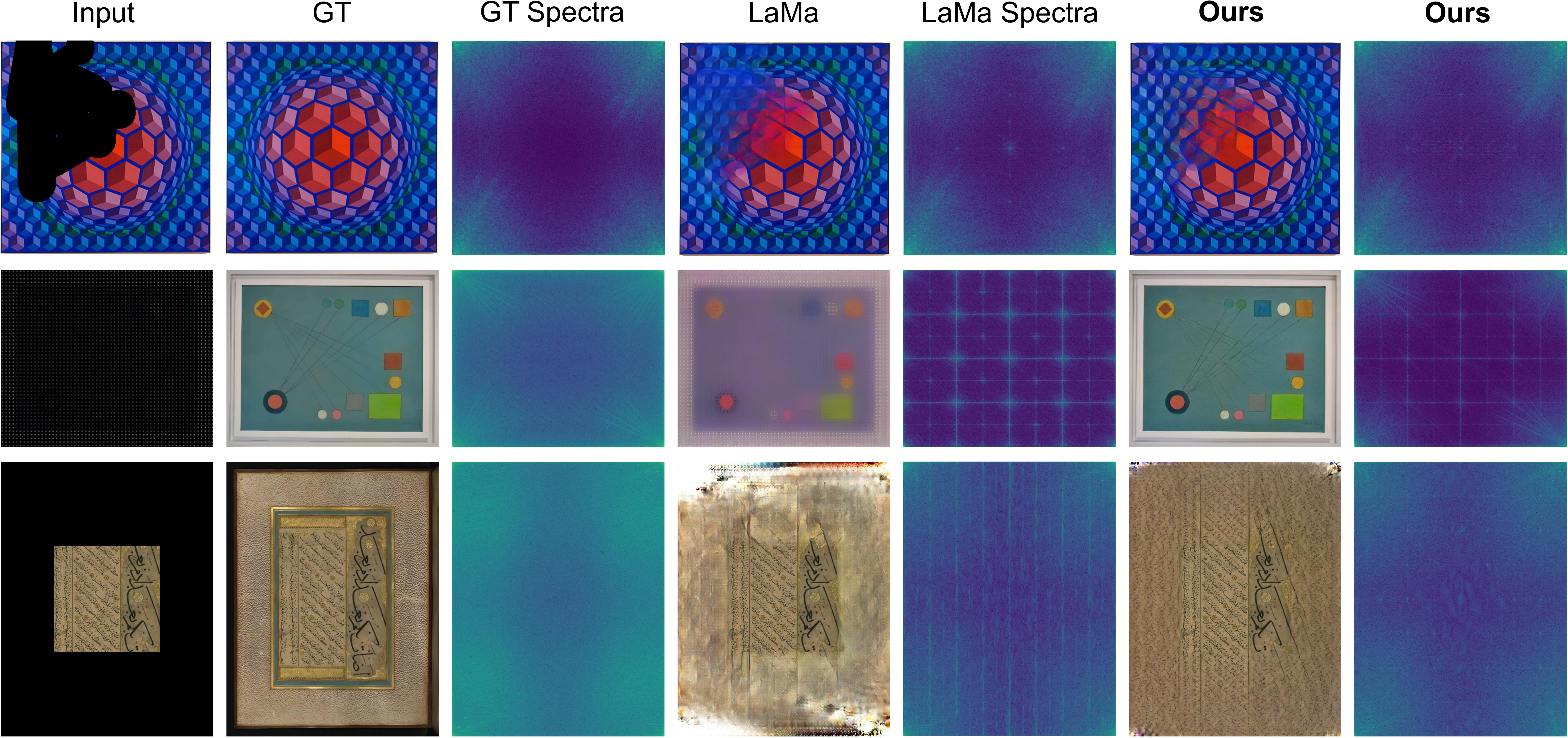}
    \figcaption{Visualization of the spectra of real and fake images. The spectra of images generated by our method is cleaner than the LaMa \cite{9707077}. At the same time, the images of our method are more realistic and cleaner.}
    \label{Figure 6}
\end{figure*}

In view of the dazzling performance of transformer architecture in other tasks, some transformer based methods have emerged recently.
For example, Wan et al. \cite{wan2021high} propose the first transformer based image inpainting method to get the image prior and send the image prior to a CNN.
To incorporate the image prior, the approach of \cite{yu2021diverse} designs a bidirectional and autoregressive transformer.
More recently, Dong et al. \cite{Dong2022IncrementalTS} design a transformer by edge auxiliaries to acquire prior and send the prior with masking positional encoding to a LaMa \cite{9707077} like network.
These methods based on transformer enjoy excellent performance compared with CNN-based methods.

In addition, there is another line of research based on diffusion model.
SohlDickstein et al. \cite{SohlDickstein2015DeepUL} firstly utilize early diffusion models for image inpainting, while Song et al. \cite{Song2021ScoreBasedGM} propose a score-based method using stochastic differential equations for image generation.
More recently, Andreas et al. \cite{DBLP:journals/corr/abs-2201-09865} propose a special model for image inpainting task which is based on diffusion model.

\subsection{Loss Functions}
In the image restoration task, the most commonly used reconstruction loss is $L_1$ loss, $L_2$ loss and Charbonnier loss.
Some image restoration methods also use perceptual loss \cite{Johnson2016PerceptualLF} and adversarial loss \cite{Goodfellow2014GenerativeAN} to improve the perceptive performance.
But above-mentioned methods all focus on the spatial domain reconstruction.
In order to improve the reconstructed image quality, optimization in the frequency domain has gradually attracted researchers' attention in recent years.
For example, spectral regularization is a preliminary attempt \cite{Liu2019SpectralRF}.
More recently, Gal et al. \cite{Gal2021SWAGANAS} propose a wavelet based image generation method.
Jiang et al. \cite{9710849} introduce the focal frequency loss which focuses on hard frequencies.
Meanwhile, there are some other work \cite{Cai2021FrequencyDI, Jung2021SpectralDA} on image restoration in the frequency domain.



\section{Our Method}
\label{sec:method}


\begin{figure*}[!t]
    \centering
    \includegraphics[width=0.98\linewidth]{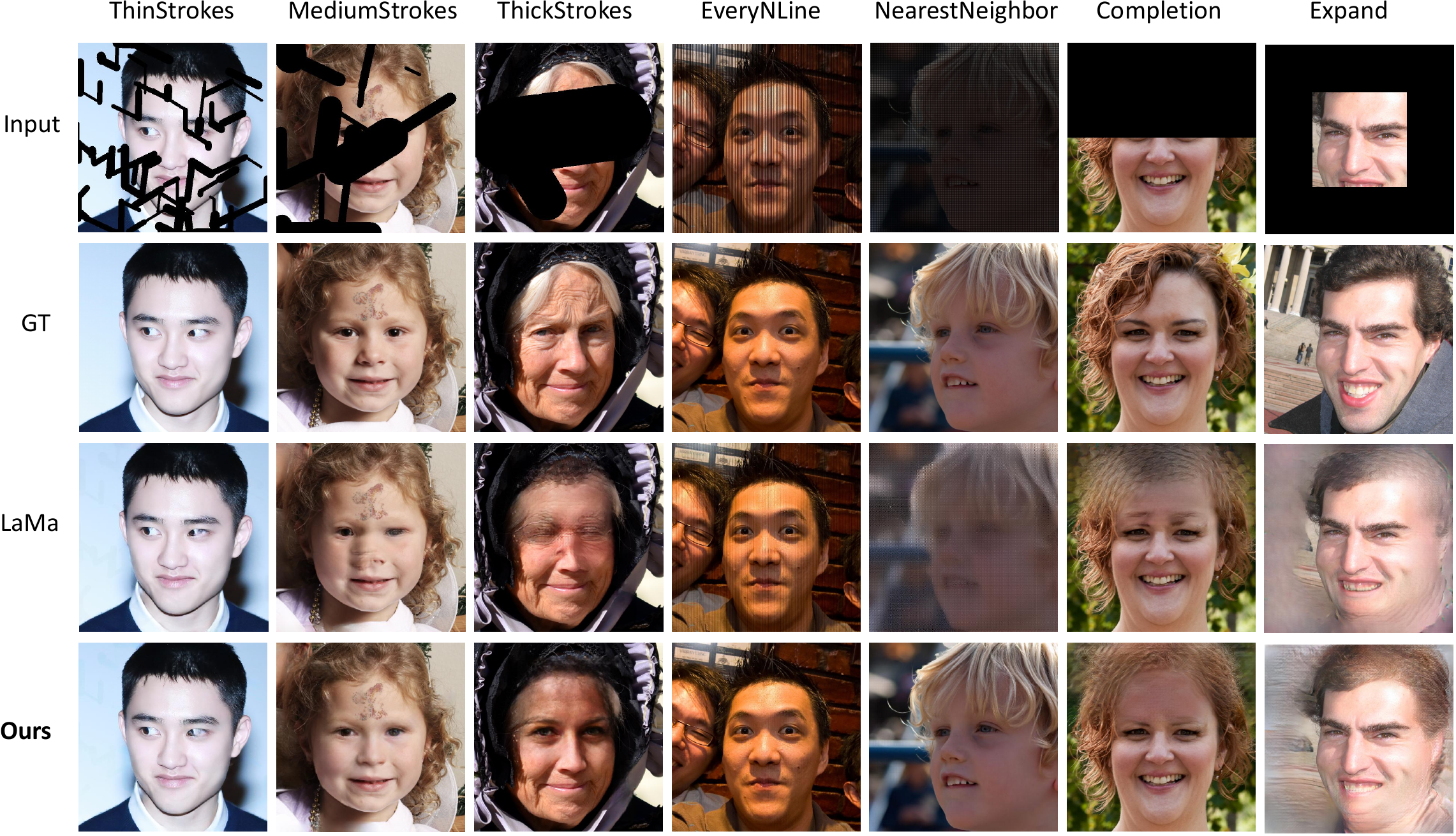}
    \figcaption{Comparison with the state-of-the-art method: LaMa \cite{9707077} for Face Inpainting over seven mask settings on FFHQ \cite{DBLP:journals/corr/abs-1812-04948} dataset. The face images generated by our method are more realistic and have more details on the hair and facial features.}
    \label{Figure 3}
\end{figure*}

\subsection{Network Architecture}
\label{sec:Network Architecture}
Fast Fourier Convolution (FFC) in introduced in \cite{Chi2020FastFC} to capture the global receptive field in the frequency domain. Inspired by this work, Suvorov et al. \cite{9707077} design a network architecture called LaMa which achieves the state-of-the-art results using FFC to capture the global information.
Following these pioneering work, we use LaMa as our network architecture.
Like some other inpainting models, LaMa \cite{9707077} uses an AutoEncoder model to extract the image features. The bottleneck module includes several FFC layers which are based on a channel-wise Fast Fourier transform (FFT), thus capturing global context information.
The FFC layer splits feature channels into two branches: the local branch uses regular convolutions to obtain local features, while the global branch leverages FFC to obtain global features.

\subsection{Training with General Mask}
\label{sec:General Mask Training Strategies}
Each training image $x$ is from a training dataset superimposed by a synthetically generated mask.
We train four different models for four datasets: FFHQ \cite{DBLP:journals/corr/abs-1812-04948}, ImageNet \cite{5206848}, Places2 \cite{zhou2017places} and WikiArt \cite{Saleh2015LargescaleCO}, respectively.
After a lot of experiments, we find that the policy of mask generation noticeably influences the performance of the inpainting model as shown in Tab. \ref{Table 8}.

We firstly try an aggressive large mask generation strategy based on LaMa \cite{9707077}.
This strategy uniformly uses samples from polygonal strip dilated by random strokes and rectangles of arbitrary aspect ratios.
However, from the experimental results shown in Tab. \ref{Table 8}, we can learn that the model trained with this mask generation strategy acquires poor results with the types of ``Completion'', ``Expand'', ``Every n line'' and ``Nearest Neighbor'' masks.

\begin{figure*}[!t]
    \centering
    \includegraphics[width=0.98\linewidth]{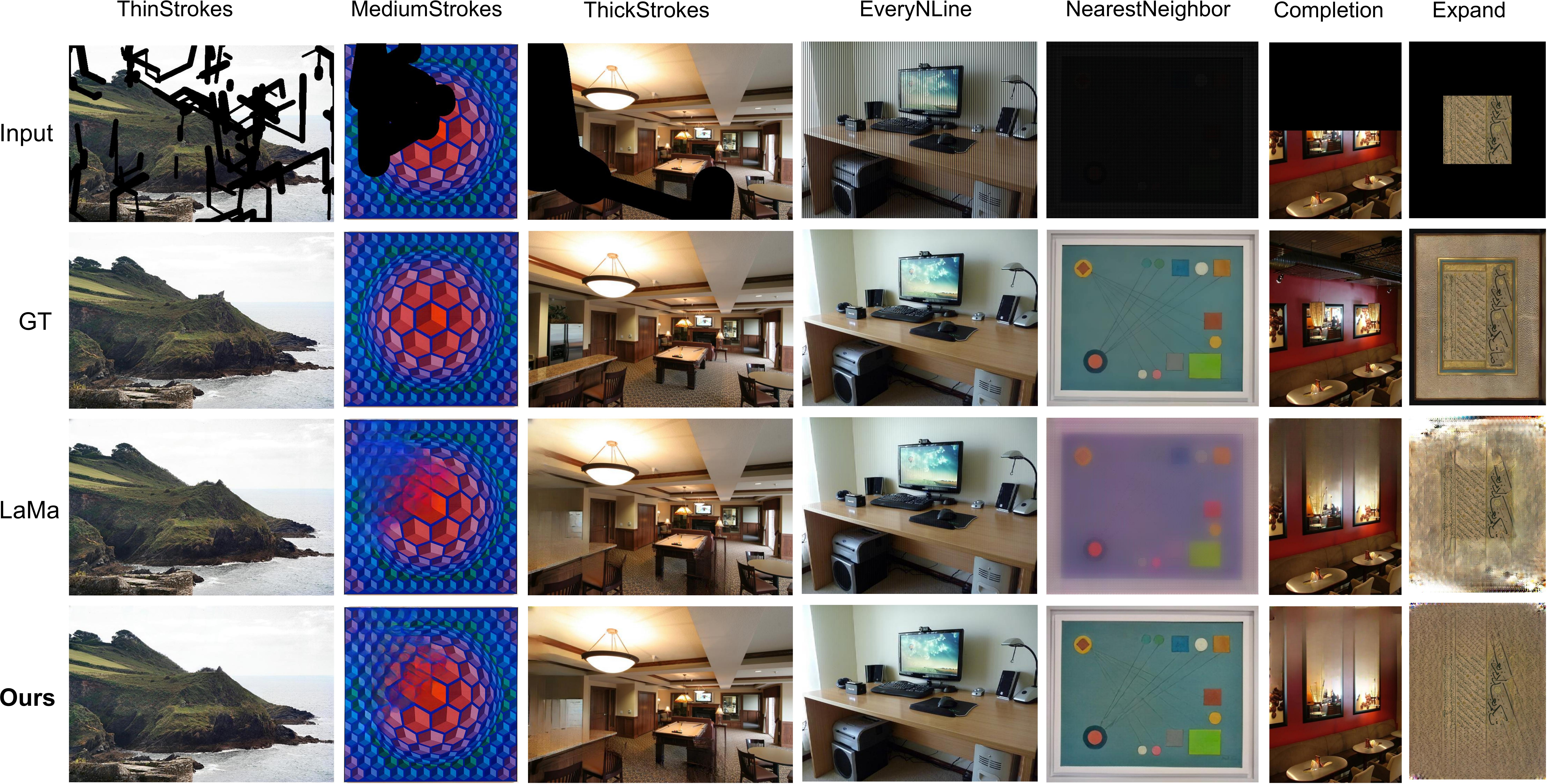}
    \figcaption{Comparison with the state-of-the-art method: LaMa \cite{9707077} for Inpainting over seven mask settings on ImageNet \cite{5206848}, Places2 \cite{zhou2017places} and WikiArt \cite{Saleh2015LargescaleCO} datasets. The fourth column is sampled from ImageNet \cite{5206848} dataset. The first, third and sixth columns are sampled from Places2 \cite{zhou2017places} dataset. The second, fifth and seventh columns are sampled from WikiArt \cite{Saleh2015LargescaleCO} dataset.}
    \label{Figure 5}
\end{figure*}

Subsequently, unlike the conventional practice, e.g. DeepFillv2 \cite{9010689} or LaMa \cite{9707077}, we utilize seven types of mask generation strategies, ``Completion'', ``Expand'', ``Every n line'', ``Nearest Neighbor'', ``Thin Strokes'', ``Medium Strokes'', and ``Thick Strokes'', to generate samples randomly, which we call it \textit{General Mask} generation strategy.
These generated masks include not only small narrow masks, but also large-area masks.
As shown in Tabs. \ref{Table 1} and \ref{Table 2}, the model trained with \textit{General Mask} generation strategy performs better than original LaMa \cite{9707077} with all seven types of mask.
This demonstrates that the diversity of degraded images can improve the robustness of the model to face various types of mask.

\subsection{Loss Functions}
\label{sec:Our Losses}
The inpainting problem is ambiguous, which means one image that needs to be repaired might correspond to multiple images. One way is introduce some constraints to alleviate this. In this paper, we will introduce a joint spatial and frequency loss to regularize the optimization of our model, which regularizes the reconstruction results from both the spatial and frequency domains.

For the image inpainting task, the loss in the spatial domain is indispensable.
First of all, we use $L_1$ loss between the unmasked regions in the spatial domain.
It can be formulated as:
\begin{equation}
\begin{aligned}
\mathcal{L}_{1}=\| \boldsymbol{x}-\boldsymbol{\hat{x}} \|_{1} \odot (1-\boldsymbol{M})
\end{aligned}
\end{equation}
where $x$ and $\hat{x}$ indicate the ground truth and predicted images respectively.
$\boldsymbol{M}$ represents 0-1 mask (1 means masked regions and 0 means no masked regions) and $\odot$ means the element-wise multiplication.
Moreover, we use the high receptive field perceptual loss $\mathcal{L}_{PL}$.
It can be formulated as:
\begin{equation}
\begin{aligned}
\mathcal{L}_{PL}=\left[\phi_{PL}(\boldsymbol{x})-\phi_{PL}(\hat{\boldsymbol{x}})\right]^{2}
\end{aligned}
\end{equation}
where $\phi$ indicates a pretrained segmentation ResNet50\cite{He2016DeepRL} with dilated convolutions.

Denote $\mathcal{L}_D$ the discriminator loss, $\mathcal{L}_G$ the generator loss and $\mathcal{L}_{P}$ the gradient penalty, the adversarial loss can be formulated as:
\begin{equation}
\begin{aligned}
\mathcal{L}_{D}=&-\mathbb{E}_{\boldsymbol{x}}[\log D(\boldsymbol{x})]-\mathbb{E}_{\hat{\boldsymbol{x}}, \boldsymbol{M}}[\log D(\hat{\boldsymbol{x}}) \odot(1-\boldsymbol{M})] \\ &-\mathbb{E}_{\hat{\boldsymbol{x}},\boldsymbol{M}}[\log (1-D(\hat{\boldsymbol{x}})) \odot \boldsymbol{M}]
\end{aligned}
\end{equation}
Notably, we only regard features from masked regions as fake samples in $\mathcal{L}_D$.

\begin{equation}
\begin{aligned}
\mathcal{L}_{G}=-\mathbb{E}_{\hat{\boldsymbol{x}}}[\log D(\hat{\boldsymbol{x}})]
\end{aligned}
\end{equation}

\begin{equation}
\begin{aligned}
\mathcal{L}_{P}=\mathbb{E}_{\boldsymbol{x}}\left\|\nabla_{\boldsymbol{x}} D(\boldsymbol{x})\right\|^{2}
\end{aligned}
\end{equation}

\begin{equation}
\begin{aligned}
\mathcal{L}_{a d v}=\mathcal{L}_{D}+\mathcal{L}_{G}+\lambda_{P} \mathcal{L}_{P}
\end{aligned}
\end{equation}
We also use the feature match loss $\mathcal{L}_{fm}$, which is based on $L_1$ loss between discriminator features of true and fake images, to stable the GAN training.
Finally, the loss of LaMa\cite{9707077} can be written as:
\begin{equation}
\begin{aligned}
\mathcal{L}_{LaMa}=\lambda_{1} \mathcal{L}_{L 1}+\lambda_{a d v} \mathcal{L}_{a d v}+\lambda_{f m} \mathcal{L}_{f m}+\lambda_{PL} \mathcal{L}_{PL}
\end{aligned}
\end{equation}
where $\lambda_{1} = 10$, $\lambda_{a d v} = 10$, $\lambda_{PL} = 100$ and $\lambda_{f m} = 30$.

We find although LaMa \cite{9707077} uses Fast Fourier Convolution, the model is not optimized in the frequency domain.
Therefore, the spectrum of the images generated by LaMa \cite{9707077} is defective, as shown in Fig. \ref{Figure 6}.
With this in mind, we take the focal frequency loss \cite{9707077} to construct the frequency fidelity term. Specifically, the focal frequency loss $\mathcal{L}_{FFL}$ is defined as follows:
\begin{equation}
\begin{aligned}
\mathcal{L}_{FFL} = \frac{1}{MN}\sum_{u=0}^{M-1}\sum_{v=0}^{N-1} w(u,v)|F_r(u,v)-F_f(u,v)|^{2}
\end{aligned}
\end{equation}
where $F_r$ represents 2D discrete Fourier transform of real image (represents $x$ here), and $F_f$ represents 2D discrete Fourier transform of fake image (represents $\hat x$ here).
The matrix $w(u,v)$ represents the weight for the spatial frequency at coordinate $(u,v)$, which is defined as:
\begin{equation}
\begin{aligned}
w(u,v) = |F_r(u,v)-F_f(u,v)|^\alpha
\end{aligned}
\end{equation}
where $\alpha = 1$.
Furthermore, we find that the reconstructed images of LaMa \cite{9707077} always get obvious checkerboard effect and ripples with the mask of ``Nearest Neighbor'' and ``Every N Line''. Based on this finding, we introduce the total variation loss, which is defined as: \cite{SohlDickstein2015DeepUL}:
\begin{equation}
\begin{aligned}
\mathcal{L}_{TV} = \sum_{i,j}((x_{i,j+1}-x_{i,j})^2 + (x_{i+1,j}-x_{i,j})^2)^{\frac{\beta}{2}}
\end{aligned}
\end{equation}
where $\beta = 2$.

The final joint spatial and frequency loss function for our inpainting model can be written as:
\begin{equation}
\begin{aligned}
\mathcal{L} = \alpha_1 \mathcal{L}_{TV} + \alpha_2 \mathcal{L}_{FFL} + \alpha_3 \mathcal{L}_{LaMa}
\end{aligned}
\end{equation}
where $\alpha_1 = 1$, $\alpha_2 = 1$ and $\alpha_3 = 1$.

\begin{table*}[ht]

\resizebox{\linewidth}{!}{
\begin{tabular}{c|c|ccccccc}
\hline
        \multirow{2}{*}{Datasets} & \multirow{2}{*}{Methods} & \multicolumn{7}{c}{LPIPS ($\downarrow$)}                      \\
                          &                         & \multicolumn{1}{c|}{Completion}      & \multicolumn{1}{c|}{Expand}          & \multicolumn{1}{c|}{NearesNeighbor} & \multicolumn{1}{c|}{ThinStrokes}     & \multicolumn{1}{c|}{EveryNLines} & \multicolumn{1}{c|}{MediumStrokes}   & ThickStrokes \\ \cline{1-9}
\multirow{3}{*}{Places2}  & LaMa                    & \multicolumn{1}{c|}{0.3302}          & \multicolumn{1}{c|}{0.6098}          & \multicolumn{1}{c|}{0.6511}            & \multicolumn{1}{c|}{0.0903}          & \multicolumn{1}{c|}{0.3125}          & \multicolumn{1}{c|}{0.1193}          & 0.1374 \\
                          & Big LaMa                & \multicolumn{1}{c|}{\color[HTML]{3166FF}0.3298}          & \multicolumn{1}{c|}{\color[HTML]{3166FF}0.6031}          & \multicolumn{1}{c|}{\color[HTML]{3166FF}0.6473}            & \multicolumn{1}{c|}{\color[HTML]{FE0000}{0.0807}} & \multicolumn{1}{c|}{\color[HTML]{3166FF}0.2788}          & \multicolumn{1}{c|}{\color[HTML]{FE0000}{0.1026}} & \color[HTML]{3166FF}0.1266 \\
                          & \textbf{GLaMa}                   & \multicolumn{1}{c|}{\color[HTML]{FE0000}{0.3212}} & \multicolumn{1}{c|}{\color[HTML]{FE0000}{0.5923}} & \multicolumn{1}{c|}{\color[HTML]{FE0000}{0.2280}}   & \multicolumn{1}{c|}{\color[HTML]{3166FF}0.0890}           & \multicolumn{1}{c|}{\color[HTML]{FE0000}{0.0891}} & \multicolumn{1}{c|}{\color[HTML]{3166FF}0.1088}          & \color[HTML]{FE0000}{0.0774}  \\ \cline{1-9}
\multirow{3}{*}{FFHQ}     & LaMa                    & \multicolumn{1}{c|}{0.3102}          & \multicolumn{1}{c|}{0.6089}          & \multicolumn{1}{c|}{0.6518}            & \multicolumn{1}{c|}{0.0976}          & \multicolumn{1}{c|}{0.312}           & \multicolumn{1}{c|}{0.1124}          & 0.1307  \\
                          & Big LaMa                & \multicolumn{1}{c|}{\color[HTML]{3166FF}0.3048}          & \multicolumn{1}{c|}{\color[HTML]{3166FF}0.5892}          & \multicolumn{1}{c|}{\color[HTML]{3166FF}0.6081}            & \multicolumn{1}{c|}{\color[HTML]{FE0000}{0.1109}} & \multicolumn{1}{c|}{\color[HTML]{3166FF}0.3067}          & \multicolumn{1}{c|}{\color[HTML]{FE0000}{0.1246}} & \color[HTML]{3166FF}0.1331  \\
                          & \textbf{GLaMa}                   & \multicolumn{1}{c|}{\color[HTML]{FE0000}{0.2932}} & \multicolumn{1}{c|}{\color[HTML]{FE0000}{0.5532}} & \multicolumn{1}{c|}{\color[HTML]{FE0000}{0.2402}}   & \multicolumn{1}{c|}{\color[HTML]{3166FF}0.1121}          & \multicolumn{1}{c|}{\color[HTML]{FE0000}{0.1829}} & \multicolumn{1}{c|}{\color[HTML]{3166FF}0.1269}          & \color[HTML]{FE0000}{0.1324}  \\ \cline{1-9}
\multirow{3}{*}{ImageNet} & LaMa                    & \multicolumn{1}{c|}{0.3106}          & \multicolumn{1}{c|}{0.5877}          & \multicolumn{1}{c|}{0.6305}            & \multicolumn{1}{c|}{0.0768}          & \multicolumn{1}{c|}{0.2946}          & \multicolumn{1}{c|}{0.0997}          & 0.1133  \\
                          & Big LaMa                & \multicolumn{1}{c|}{\color[HTML]{3166FF}0.3102}          & \multicolumn{1}{c|}{\color[HTML]{3166FF}0.5834}          & \multicolumn{1}{c|}{\color[HTML]{3166FF}0.6283}            & \multicolumn{1}{c|}{\color[HTML]{FE0000}{0.0607}} & \multicolumn{1}{c|}{\color[HTML]{3166FF}0.2529}          & \multicolumn{1}{c|}{\color[HTML]{FE0000}{0.0815}} & \color[HTML]{3166FF}0.1010 \\
                          & \textbf{GLaMa}                   & \multicolumn{1}{c|}{\color[HTML]{FE0000}{0.3009}} & \multicolumn{1}{c|}{\color[HTML]{FE0000}{0.5719}} & \multicolumn{1}{c|}{\color[HTML]{FE0000}{0.2061}}   & \multicolumn{1}{c|}{\color[HTML]{3166FF}0.0654}          & \multicolumn{1}{c|}{\color[HTML]{FE0000}{0.0651}} & \multicolumn{1}{c|}{\color[HTML]{3166FF}0.0834}          & \color[HTML]{FE0000}{0.0859} \\ \cline{1-9}
\multirow{3}{*}{WikiArt}  & LaMa                    & \multicolumn{1}{c|}{0.3517}          & \multicolumn{1}{c|}{0.6261}          & \multicolumn{1}{c|}{0.6711}            & \multicolumn{1}{c|}{0.1093}          & \multicolumn{1}{c|}{0.3317}          & \multicolumn{1}{c|}{0.1397}          & 0.1582 \\
                          & Big LaMa                & \multicolumn{1}{c|}{\color[HTML]{3166FF}0.3498}          & \multicolumn{1}{c|}{\color[HTML]{3166FF}0.6217}          & \multicolumn{1}{c|}{\color[HTML]{3166FF}0.6673}            & \multicolumn{1}{c|}{\color[HTML]{FE0000}{0.1008}} & \multicolumn{1}{c|}{\color[HTML]{3166FF}0.2889}          & \multicolumn{1}{c|}{\color[HTML]{FE0000}{0.1219}} & \color[HTML]{3166FF}0.1439  \\
                          & \textbf{GLaMa}                  & \multicolumn{1}{c|}{\color[HTML]{FE0000}{0.3311}} & \multicolumn{1}{c|}{\color[HTML]{FE0000}{0.5989}} & \multicolumn{1}{c|}{\color[HTML]{FE0000}{0.2497}}   & \multicolumn{1}{c|}{\color[HTML]{3166FF}0.1065}          & \multicolumn{1}{c|}{\color[HTML]{FE0000}{0.0870}} & \multicolumn{1}{c|}{\color[HTML]{3166FF}0.1244}          & \color[HTML]{FE0000}{0.0932} \\
                          \hline
    \end{tabular}
}
\vspace{-0.2cm}
\caption{LPIPS comparisons of different methods on the four datasets.
The top two performing method is highlighted in {\color[HTML]{FE0000}red} and {\color[HTML]{3166FF}blue}.
}
    \label{Table 1}
\end{table*}

\begin{table*}[ht]

\resizebox{\linewidth}{!}{
\begin{tabular}{c|c|ccccccc}
\hline
\multirow{2}{*}{Datasets} & \multirow{2}{*}{Methods} & \multicolumn{7}{c}{FID ($\downarrow$)}                               \\
                          &                         & \multicolumn{1}{c|}{Completion}     & \multicolumn{1}{c|}{Expand}         & \multicolumn{1}{c|}{NearestNeighbor} & \multicolumn{1}{c|}{ThinStrokes}    & \multicolumn{1}{c|}{EveryNLines} & \multicolumn{1}{c|}{MediumStrokes}  & ThickStrokes \\ \cline{1-9}
\multirow{3}{*}{Places2}  & LaMa                    & \multicolumn{1}{c|}{33.63}          & \multicolumn{1}{c|}{87.35}          & \multicolumn{1}{c|}{163.92}            & \multicolumn{1}{c|}{6.37}           & \multicolumn{1}{c|}{17.07}           & \multicolumn{1}{c|}{10.17}          & 13.21 \\
                          & Big LaMa                & \multicolumn{1}{c|}{\color[HTML]{3166FF}30.76}          & \multicolumn{1}{c|}{\color[HTML]{3166FF}81.73}          & \multicolumn{1}{c|}{\color[HTML]{3166FF}167.31}            & \multicolumn{1}{c|}{\color[HTML]{FE0000}{5.74}}  & \multicolumn{1}{c|}{\color[HTML]{3166FF}14.04}           & \multicolumn{1}{c|}{\color[HTML]{FE0000}{9.18}}  & \color[HTML]{FE0000}{12.08} \\ \
                          & \textbf{GLaMa}                   & \multicolumn{1}{c|}{\color[HTML]{FE0000}{29.98}} & \multicolumn{1}{c|}{\color[HTML]{FE0000}{68.74}} & \multicolumn{1}{c|}{\color[HTML]{FE0000}{8.16}}     & \multicolumn{1}{c|}{\color[HTML]{3166FF}6.29}           & \multicolumn{1}{c|}{\color[HTML]{FE0000}{2.05}}   & \multicolumn{1}{c|}{\color[HTML]{3166FF}10.02}          & \color[HTML]{3166FF}13.03 \\ \cline{1-9}
\multirow{3}{*}{FFHQ}     & LaMa                    & \multicolumn{1}{c|}{23.82}          & \multicolumn{1}{c|}{117.60}         & \multicolumn{1}{c|}{133.70}            & \multicolumn{1}{c|}{6.60}           & \multicolumn{1}{c|}{21.44}           & \multicolumn{1}{c|}{8.64}           & 9.19 \\
                          & Big LaMa                & \multicolumn{1}{c|}{\color[HTML]{3166FF}23.32}          & \multicolumn{1}{c|}{\color[HTML]{3166FF}111.91}         & \multicolumn{1}{c|}{\color[HTML]{3166FF}131.72}            & \multicolumn{1}{c|}{\color[HTML]{FE0000}{5.89}}  & \multicolumn{1}{c|}{\color[HTML]{3166FF}20.01}           & \multicolumn{1}{c|}{\color[HTML]{FE0000}{6.68}}  & \color[HTML]{FE0000}{7.54} \\
                          & \textbf{GLaMa}                  & \multicolumn{1}{c|}{\color[HTML]{FE0000}{20.37}} & \multicolumn{1}{c|}{\color[HTML]{FE0000}{89.70}} & \multicolumn{1}{c|}{\color[HTML]{FE0000}{7.65}}     & \multicolumn{1}{c|}{\color[HTML]{3166FF}6.10}           & \multicolumn{1}{c|}{\color[HTML]{FE0000}{6.21}}   & \multicolumn{1}{c|}{\color[HTML]{3166FF}7.65}           & \color[HTML]{3166FF}8.09 \\ \cline{1-9}
\multirow{3}{*}{ImageNet} & LaMa                    & \multicolumn{1}{c|}{25.28}          & \multicolumn{1}{c|}{119.34}         & \multicolumn{1}{c|}{135.01}            & \multicolumn{1}{c|}{8.69}           & \multicolumn{1}{c|}{23.90}           & \multicolumn{1}{c|}{10.53}          & 11.19  \\
                          & Big LaMa                & \multicolumn{1}{c|}{\color[HTML]{3166FF}25.31}          & \multicolumn{1}{c|}{\color[HTML]{3166FF}110.78}         & \multicolumn{1}{c|}{\color[HTML]{3166FF}131.47}            & \multicolumn{1}{c|}{\color[HTML]{FE0000}{7.53}}  & \multicolumn{1}{c|}{\color[HTML]{3166FF}22.17}           & \multicolumn{1}{c|}{\color[HTML]{FE0000}{8.77}}  & \color[HTML]{FE0000}{9.98} \\
                          & \textbf{GLaMa}                  & \multicolumn{1}{c|}{\color[HTML]{FE0000}{22.91}} & \multicolumn{1}{c|}{\color[HTML]{FE0000}{91.78}} & \multicolumn{1}{c|}{\color[HTML]{FE0000}{9.78}}     & \multicolumn{1}{c|}{\color[HTML]{3166FF}8.06}           & \multicolumn{1}{c|}{\color[HTML]{FE0000}{8.21}}   & \multicolumn{1}{c|}{\color[HTML]{3166FF}9.24}           & \color[HTML]{3166FF}10.01 \\ \cline{1-9}
\multirow{3}{*}{WikiArt}  & LaMa                    & \multicolumn{1}{c|}{39.13}          & \multicolumn{1}{c|}{92.01}          & \multicolumn{1}{c|}{162.59}            & \multicolumn{1}{c|}{12.88}          & \multicolumn{1}{c|}{23.47}           & \multicolumn{1}{c|}{16.75}          & 18.96 \\
                          & Big LaMa                & \multicolumn{1}{c|}{\color[HTML]{3166FF}36.76}          & \multicolumn{1}{c|}{\color[HTML]{3166FF}86.18}          & \multicolumn{1}{c|}{\color[HTML]{3166FF}153.31}            & \multicolumn{1}{c|}{\color[HTML]{FE0000}{11.18}} & \multicolumn{1}{c|}{\color[HTML]{3166FF}20.62}           & \multicolumn{1}{c|}{\color[HTML]{FE0000}{15.81}} & \color[HTML]{FE0000}{17.04} \\
                          & \textbf{GLaMa}                  & \multicolumn{1}{c|}{\color[HTML]{FE0000}{35.72}} & \multicolumn{1}{c|}{\color[HTML]{FE0000}{74.51}} & \multicolumn{1}{c|}{\color[HTML]{FE0000}{12.08}}    & \multicolumn{1}{c|}{\color[HTML]{3166FF}11.89}          & \multicolumn{1}{c|}{\color[HTML]{FE0000}{8.03}}   & \multicolumn{1}{c|}{\color[HTML]{3166FF}16.23}          & \color[HTML]{3166FF}17.87 \\
                          \hline
    \end{tabular}
}
\vspace{-0.2cm}
\caption{
FID comparisons of different methods on the four datasets.}
\label{Table 2}
\end{table*}

\section{Experiments}
\label{sec:Experiments}

\subsection{Datasets and Metrics}
\label{sec:Datasets and metrics}
In our experiments, we use FFHQ \cite{DBLP:journals/corr/abs-1812-04948}, ImageNet \cite{5206848}, Places2 \cite{zhou2017places} and WikiArt \cite{Saleh2015LargescaleCO} datasets, which contain various kinds of images, to demonstrate the results.
For FFHQ \cite{DBLP:journals/corr/abs-1812-04948}, we use about 60K face images as the training set and 2,000 other images as the validation.
For Places2 \cite{zhou2017places}, we use about 1,800K images from various scenes as the training set and 3650 other images from validation set as the validation.
For ImageNet \cite{5206848}, we use about 1,000K images as the training set and 2000 other images from validation dataset as the validation set.
And for WikiArt \cite{Saleh2015LargescaleCO}, we use the whole training set and 2000 other images from validation set as the validation.

In the training stage for FFHQ \cite{DBLP:journals/corr/abs-1812-04948}, our images are resized to $256 \times 256$ resolution.
For the remaining three training datasets, we crop the images to $256 \times 256$ pixels.
Moreover, we use the original size of the images in the validation stage without cropping or resizing them.

We follow the established practice in recent image inpainting literature and use Learned Perceptual Image Patch Similarity (LPIPS) \cite{8578166} and Frechet inception distance (FID) \cite{Heusel2017GANsTB} metrics.
Compared to PSNR and SSIM \cite{1284395}, LPIPS \cite{8578166} and FID \cite{Heusel2017GANsTB} are more suitable for measuring performance of inpainting for large masks.

\subsection{Implementation Details}

\label{sec:Implementation Details}
For inpainting network \cite{9707077}, we followed LaMa \cite{9707077} using a ResNet-like \cite{He2016DeepRL} architecture with 3 downsampling blocks, 6-18 residual blocks, and 3 upsampling blocks.
We use Adam optimizer \cite{Kingma2015AdamAM}, with the fixed learning rates 0.001 and 0.0001 for inpainting and discriminator networks, respectively.
We set the hyper-parameters using the grid search strategy on FFHQ dataset, which leds to the weight values $\alpha_1= 1$, $\alpha_2=1$, $\alpha_3=1$, $\alpha=1$, $\beta=2$.
The same parameters are used for other three datasets for training.
All models are trained for 40 epochs with a batch size of 20, on 8 NVidia V100 GPUs for approximately 72 hours.
The LaMa and Big LaMa \cite{9707077} models used as baseline in the Tabs. \ref{Table 1} and \ref{Table 2} were retrained with the same experimental setups. Becaused LaMa and Big LaMa \cite{9707077} and the results are higher than the original LaMa and Big LaMa \cite{9707077} model in FFHQ \cite{DBLP:journals/corr/abs-1812-04948} and Places2 \cite{zhou2017places} datasets.

\subsection{Comparisons to the Baselines}
\label{sec:Comparisons to the baselines}
We compare the proposed approach with the strong baselines (LaMa and Big LaMa \cite{9707077}). For each dataset, we validate the performance across seven types of mask.
As shown in Tabs. \ref{Table 1} and \ref{Table 2}, we can see that our GLaMa is better than the original LaMa \cite{9707077} especially for ``Completion'' mask, ``Expand'' mask, ``Nearest Neighbor'' mask and ``Every N Lines'' mask.
Compared with the images generated by LaMa, our images are more realistic with less checkerboard effect and ripples, as shown in Fig. \ref{Figure 6}, \ref{Figure 3} and \ref{Figure 5}.
GLaMa also works better than Big LaMa \cite{9707077} in most types of mask with the same network architecture and training time as LaMa \cite{9707077}.
Meanwhile, our method that simply using the \emph{General Mask} generation strategy and joint spatial and frequency loss can bring considerable improvements.

\begin{table}[ht]
\centering
\resizebox{0.75\linewidth}{!}{
\begin{tabular}{ccc|c|c}
\hline
\multicolumn{1}{c}{{ $\mathcal{L}_{LaMa}$}}  & \multicolumn{1}{c}{{$\mathcal{L}_{TV}$}} &$\mathcal{L}_{FFL}$ & {FID ($\downarrow$)}         & {LPIPS ($\downarrow$)}       \\ \hline
{\checkmark }                                                            & { }                        &    & {57.015} & {0.3382}      \\
{\checkmark }                                                             & {\checkmark}                        &    & {22.363} & {0.2350} \\
{\checkmark }                                                            & {\checkmark }                        & {\checkmark }    & {\color[HTML]{FE0000}20.827} & {\color[HTML]{FE0000}0.2253} \\
\hline
\end{tabular}
}
\caption{
Ablation studies over the loss functions.
Here, FID and LPIPS values of the results are reported.
}
\label{Table 3}
\end{table}

\begin{table}[ht]
\resizebox{\linewidth}{!}{
\begin{tabular}{ccc|c|c}
\hline
\multicolumn{1}{c}{\textit{LaMa Mask}} & \multicolumn{1}{c}{{\textit{LaMa Plus Mask}}} & \multicolumn{1}{c|}{{\textit{General Mask}}}  & {FID ($\downarrow$)}         & {LPIPS ($\downarrow$)}       \\ \hline
{\checkmark }                           & { }                                 & { }                        &{49.015} & {0.3118}      \\
{}                           & {\checkmark }                                  & { }                        &{23.514} & {0.2448} \\
{}                           & {}                                  & {\checkmark}                        &{\color[HTML]{FE0000}20.827} & {\color[HTML]{FE0000}0.2253} \\
\hline
\end{tabular}
}
\caption{Ablation studies over the mask generation methods.
\textit{LaMa Mask}, \textit{LaMa Plus Mask} represent using original LaMa, LaMa with ``Nearest Neighbour'' and ``Every n line'', respectively.
\textit{General Mask} represents using our method.
}
\label{Table 8}
\end{table}

\subsection{Ablation Studies}
\label{sec:Ablation Studies}
The goal of the ablation studies is to carefully validate the influence of different components of our method.
To investigate the effectiveness of ours mask generation strategy and losses, we conduct ablation experiments and the results comparison are shown in Tabs. \ref{Table 3} and \ref{Table 8}. All these results are conducted on FFHQ \cite{DBLP:journals/corr/abs-1812-04948} dataset.

\paragraph{Mask Generation Strategy.}
We verify that the \textit{General Mask} generation strategy is necessary in Tab. \ref{Table 8}.
\textit{LaMa Mask} and \textit{LaMa Plus Mask} generation strategy represent using LaMa original policy to generate mask (three types of mask) and LaMa original policy with ``Nearest Neighbour'' and ``Every n line'' to generate mask (five types of mask), respectively.
\textit{General Mask} generation strategy represents using our policy to generate mask (seven types of mask).
We can see from Tab. \ref{Table 8} that using \textit{General Mask} generation strategy can greatly improve the model performance.
This demonstrates that the diversity of degraded images can improve the robustness of the model to face various types of mask.

\paragraph{Loss Functions.}
We experiment with the two proposed losses in Tab. \ref{Table 3}, where total variation loss \cite{7299155} and focal frequency loss \cite{9710849} can boost the model performance.
At the same time, it can be seen from Fig. \ref{Figure 6} that the spectrum of the image generated by GLaMa is more accurate than LaMa \cite{9707077} :
ours spectrum is clean, while the spectrum of LaMa is very noisy.
These experimental results further prove the superiority of our method.

\begin{figure}
    \centering
    \includegraphics[width=0.98\linewidth]{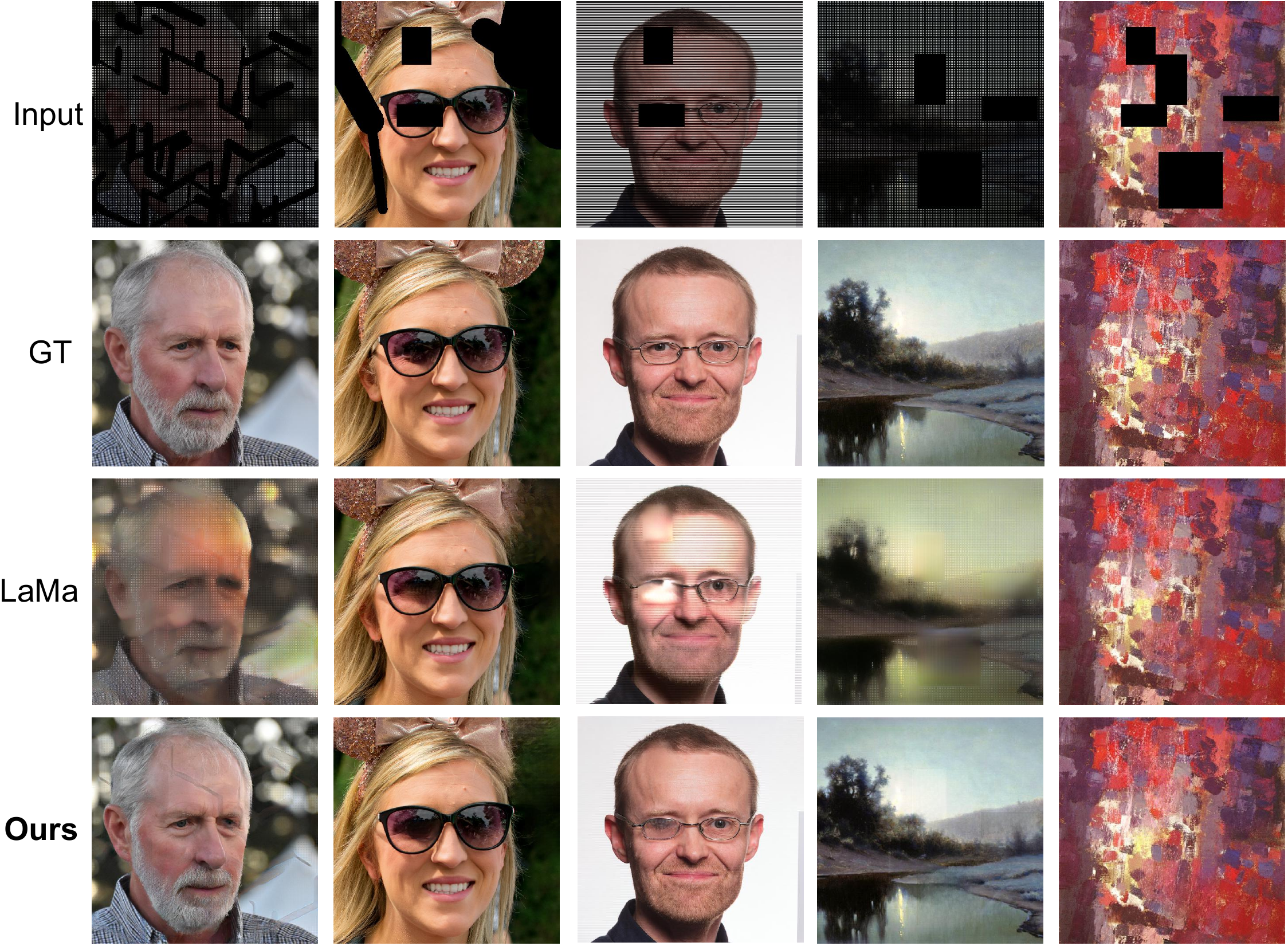}
    \caption{Visual comparisons with unseen masks.
    }
    \label{Figure 8}
\end{figure}

\subsection{Generalization of the Proposed Method}
\label{sec:Generalization}
\begin{figure*}[!t]
    \centering
    \includegraphics[width=0.905\linewidth]{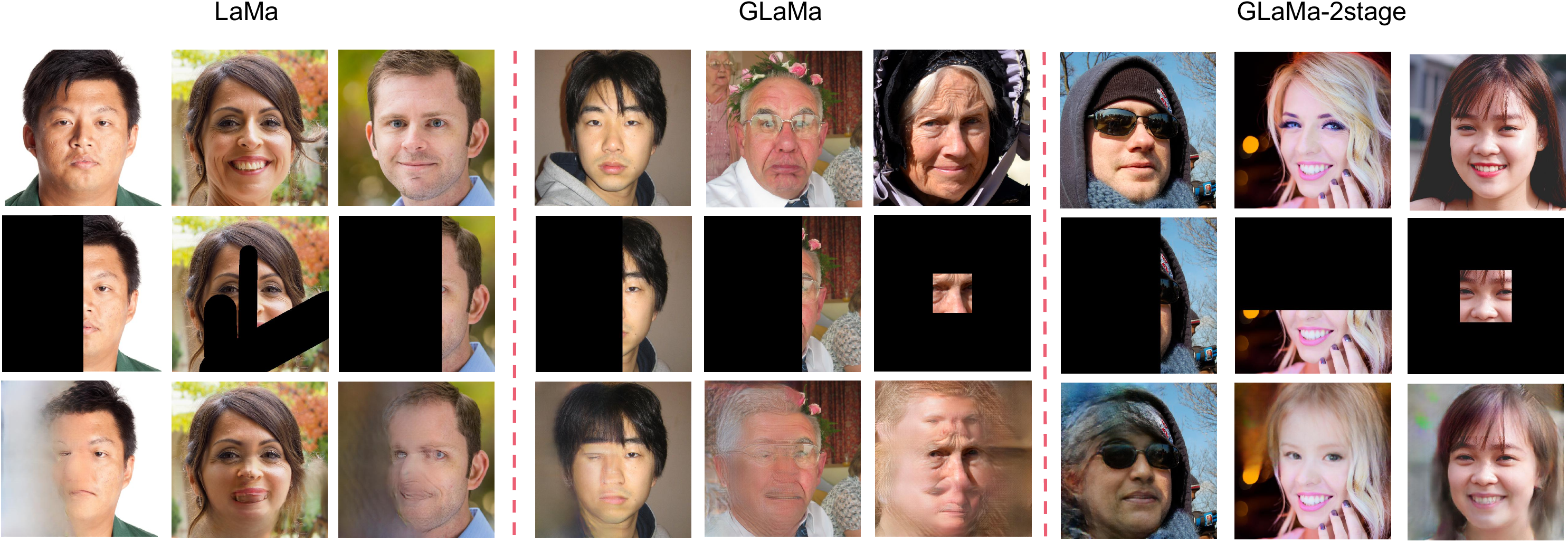}
    \caption{
    Some failed examples on the FFHQ \cite{DBLP:journals/corr/abs-1812-04948} dataset.
    }
    \label{Figure 7}
\end{figure*}

\paragraph{Generalization to the Masks.}
To verify the generalization ability of our model for the masks that do not appear in the training stage, we randomly generate some other kinds of masks that are not in the seven kinds of masks.
The mask generation strategy will combine the samples from seven kinds of masks and rectangular mask of arbitrary aspect ratios to generate some other kinds of masks.
As shown in Fig. \ref{Figure 8}, GLaMa can generate high-quality images, while the images generated by LaMa \cite{9707077} are low-quality, which proves that our method has strong generalization performance.
Hence one can see that, whether the masks are appeared during the training stage, our model can recover the target images very well.

\begin{center}
\begin{table}
\resizebox{0.99\linewidth}{!}{
\begin{tabular}{c|ccccccc|ccccccc}
\hline
\multirow{2}{*}{Teams}
                &
\multicolumn{7}{c}{FFHQ}                                                                                                                                                         \\ \cline{2-8}
                 & \multicolumn{1}{c|}{FID ($\downarrow$)}                          & \multicolumn{2}{c|}{LPIPS ($\downarrow$)}                        & \multicolumn{2}{c|}{PSNR ($\uparrow$)}                          & \multicolumn{2}{c}{SSIM ($\uparrow$)} \\ \cline{1-8}
\multicolumn{1}{c|}{VIP}                  & \multicolumn{1}{c|}{115.922}                      & \multicolumn{2}{c|}{0.433}                        & \multicolumn{2}{c|}{17.144}                        & \multicolumn{2}{c}{0.649} \\
\multicolumn{1}{c|}{\textbf{AIIA (Ours)}}                 & \multicolumn{1}{c|}{9.823}                        & \multicolumn{2}{c|}{0.239}                        & \multicolumn{2}{c|}{{\color[HTML]{3166FF} 25.316}} & \multicolumn{2}{c}{0.814} \\
\multicolumn{1}{c|}{HSSLAB}               & \multicolumn{1}{c|}{13.504}                       & \multicolumn{2}{c|}{{\color[HTML]{3166FF} 0.236}} & \multicolumn{2}{c|}{25.187}                        & \multicolumn{2}{c}{{\color[HTML]{3166FF} 0.821}} \\
\multicolumn{1}{c|}{KwaiInpainting}       & \multicolumn{1}{c|}{21.345}                       & \multicolumn{2}{c|}{0.239}                        & \multicolumn{2}{c|}{25.060}                        & \multicolumn{2}{c}{{\color[HTML]{FE0000} 0.838}} \\
\multicolumn{1}{c|}{ArtificiallyInspired} & \multicolumn{1}{c|}{{\color[HTML]{FE0000} 4.719}} & \multicolumn{2}{c|}{{\color[HTML]{FE0000} 0.205}} & \multicolumn{2}{c|}{{\color[HTML]{FE0000} 25.999}} & \multicolumn{2}{c}{0.816} \\
\multicolumn{1}{c|}{SIGMA}                & \multicolumn{1}{c|}{{\color[HTML]{3166FF} 7.203}} & \multicolumn{2}{c|}{0.248}                        & \multicolumn{2}{c|}{24.860}                        & \multicolumn{2}{c}{0.795} \\
\hline
\hline
\multirow{2}{*}{Teams}
                &
\multicolumn{7}{c}{Places2}                                                                                                                                         \\ \cline{2-8}
                 & \multicolumn{1}{c|}{FID ($\downarrow$)}                          & \multicolumn{2}{c|}{LPIPS ($\downarrow$)}                        & \multicolumn{2}{c|}{PSNR ($\uparrow$)}                          & \multicolumn{2}{c}{SSIM ($\uparrow$)} \\ \cline{1-8}
\multicolumn{1}{c|}{VIP}                  & \multicolumn{1}{c|}{{\color[HTML]{000000} 52.471}} & \multicolumn{2}{c|}{{\color[HTML]{000000} 0.415}} & \multicolumn{2}{c|}{{\color[HTML]{000000} 17.256}} & \multicolumn{2}{c}{{\color[HTML]{000000} 0.626}} \\
\multicolumn{1}{c|}{\textbf{AIIA (Ours)}}                 & \multicolumn{1}{c|}{{\color[HTML]{3166FF} 8.772}}  & \multicolumn{2}{c|}{{\color[HTML]{FE0000} 0.224}} & \multicolumn{2}{c|}{{\color[HTML]{3166FF} 24.145}} & \multicolumn{2}{c}{{\color[HTML]{FE0000} 0.800}} \\
\multicolumn{1}{c|}{HSSLAB}               & \multicolumn{1}{c|}{{\color[HTML]{000000} 9.861}}  & \multicolumn{2}{c|}{{\color[HTML]{000000} 0.227}} & \multicolumn{2}{c|}{{\color[HTML]{FE0000} 24.345}} & \multicolumn{2}{c}{{\color[HTML]{3166FF} 0.798}} \\
\multicolumn{1}{c|}{KwaiInpainting}       & \multicolumn{1}{c|}{{\color[HTML]{000000} 18.334}} & \multicolumn{2}{c|}{{\color[HTML]{333333} 0.255}} & \multicolumn{2}{c|}{{\color[HTML]{000000} 23.410}} & \multicolumn{2}{c}{{\color[HTML]{000000} 0.787}} \\
\multicolumn{1}{c|}{ArtificiallyInspired} & \multicolumn{1}{c|}{{\color[HTML]{FE0000} 7.544}}  & \multicolumn{2}{c|}{{\color[HTML]{3166FF} 0.225}} & \multicolumn{2}{c|}{{\color[HTML]{000000} 23.248}} & \multicolumn{2}{c}{{\color[HTML]{000000} 0.777}} \\
\multicolumn{1}{c|}{SIGMA}                & \multicolumn{1}{c|}{{\color[HTML]{000000} 11.496}} & \multicolumn{2}{c|}{{\color[HTML]{000000} 0.270}} & \multicolumn{2}{c|}{{\color[HTML]{000000} 22.562}} & \multicolumn{2}{c}{{\color[HTML]{000000} 0.748}} \\
\hline
\hline
\multirow{2}{*}{Teams}
                &
\multicolumn{7}{c}{ImageNet}                                                                                                                                        \\ \cline{2-8}
                 & \multicolumn{1}{c|}{FID ($\downarrow$)}                          & \multicolumn{2}{c|}{LPIPS ($\downarrow$)}                        & \multicolumn{2}{c|}{PSNR ($\uparrow$)}                          & \multicolumn{2}{c}{SSIM ($\uparrow$)} \\ \cline{1-8}
\multicolumn{1}{c|}{VIP}                  & \multicolumn{1}{c|}{{\color[HTML]{000000} 50.898}} & \multicolumn{2}{c|}{{\color[HTML]{000000} 0.403}} & \multicolumn{2}{c|}{{\color[HTML]{000000} 17.450}} & \multicolumn{2}{c}{{\color[HTML]{000000} 0.626}} \\
\multicolumn{1}{c|}{\textbf{AIIA (Ours)}}                 & \multicolumn{1}{c|}{{\color[HTML]{FE0000} 10.007}} & \multicolumn{2}{c|}{{\color[HTML]{FE0000} 0.207}} & \multicolumn{2}{c|}{{\color[HTML]{FE0000} 25.226}} & \multicolumn{2}{c}{{\color[HTML]{FE0000} 0.800}} \\
\multicolumn{1}{c|}{HSSLAB}               & \multicolumn{1}{c|}{{\color[HTML]{3166FF} 11.770}} & \multicolumn{2}{c|}{{\color[HTML]{000000} 0.227}} & \multicolumn{2}{c|}{{\color[HTML]{3166FF} 24.303}} & \multicolumn{2}{c}{{\color[HTML]{3166FF} 0.798}} \\
\multicolumn{1}{c|}{KwaiInpainting}       & \multicolumn{1}{c|}{{\color[HTML]{000000} 18.854}} & \multicolumn{2}{c|}{{\color[HTML]{333333} 0.249}} & \multicolumn{2}{c|}{{\color[HTML]{000000} 23.804}} & \multicolumn{2}{c}{{\color[HTML]{000000} 0.787}} \\
\multicolumn{1}{c|}{ArtificiallyInspired} & \multicolumn{1}{c|}{{\color[HTML]{333333} 12.059}} & \multicolumn{2}{c|}{{\color[HTML]{3166FF} 0.217}} & \multicolumn{2}{c|}{{\color[HTML]{000000} 24.278}} & \multicolumn{2}{c}{{\color[HTML]{000000} 0.777}} \\
\multicolumn{1}{c|}{SIGMA}                & \multicolumn{1}{c|}{{\color[HTML]{000000} 19.646}} & \multicolumn{2}{c|}{{\color[HTML]{000000} 0.311}} & \multicolumn{2}{c|}{{\color[HTML]{000000} 22.454}} & \multicolumn{2}{c}{{\color[HTML]{000000} 0.748}} \\
\hline
\hline
\multirow{2}{*}{Teams}
                &
\multicolumn{7}{c}{WikiArt}                                                                                                            \\ \cline{2-8}
                 & \multicolumn{1}{c|}{FID ($\downarrow$)}                          & \multicolumn{2}{c|}{LPIPS ($\downarrow$)}                        & \multicolumn{2}{c|}{PSNR ($\uparrow$)}                          & \multicolumn{2}{c}{SSIM ($\uparrow$)} \\ \cline{1-8}
\multicolumn{1}{c|}{VIP}                  & \multicolumn{1}{c|}{{\color[HTML]{000000} 75.645}} & \multicolumn{2}{c|}{{\color[HTML]{000000} 0.437}} & \multicolumn{2}{c|}{{\color[HTML]{000000} 17.243}} & \multicolumn{2}{c}{{\color[HTML]{000000} 0.609}} \\
\multicolumn{1}{c|}{\textbf{AIIA (Ours)}}                 & \multicolumn{1}{c|}{{\color[HTML]{333333} 14.974}} & \multicolumn{2}{c|}{{\color[HTML]{FE0000} 0.244}} & \multicolumn{2}{c|}{{\color[HTML]{FE0000} 24.350}} & \multicolumn{2}{c}{{\color[HTML]{FE0000} 0.767}} \\
\multicolumn{1}{c|}{HSSLAB}               & \multicolumn{1}{c|}{{\color[HTML]{333333} 14.986}} & \multicolumn{2}{c|}{{\color[HTML]{000000} 0.254}} & \multicolumn{2}{c|}{{\color[HTML]{3166FF} 24.257}} & \multicolumn{2}{c}{{\color[HTML]{333333} 0.752}} \\
\multicolumn{1}{c|}{KwaiInpainting}       & \multicolumn{1}{r|}{{\color[HTML]{000000} 26.395}} & \multicolumn{2}{c|}{{\color[HTML]{333333} 0.276}} & \multicolumn{2}{c|}{{\color[HTML]{000000} 23.142}} & \multicolumn{2}{c}{{\color[HTML]{3166FF} 0.759}} \\
\multicolumn{1}{c|}{ArtificiallyInspired} & \multicolumn{1}{c|}{{\color[HTML]{FE0000} 8.524}}  & \multicolumn{2}{c|}{{\color[HTML]{3166FF} 0.248}} & \multicolumn{2}{c|}{{\color[HTML]{000000} 23.799}} & \multicolumn{2}{c}{{\color[HTML]{000000} 0.758}} \\
\multicolumn{1}{c|}{SIGMA}                & \multicolumn{1}{c|}{{\color[HTML]{3166FF} 14.125}} & \multicolumn{2}{c|}{{\color[HTML]{000000} 0.286}} & \multicolumn{2}{c|}{{\color[HTML]{000000} 22.717}} & \multicolumn{2}{c}{{\color[HTML]{000000} 0.720}} \\
\hline
\end{tabular}
}
\caption{Performance comparisons of different methods on the four datasets. Results are provided from the NTIRE 2022 Image Inpainting challenge report \cite{romero2022ntire}.}
\label{Table 5}
\end{table}
\end{center}

\paragraph{Generalization to the Resolution.}
To further exploit the reconstruction ability of different approaches, we also conduct some experiments when the input images are with $512 \times 512$ pixels (or higher resolution) while the training images are with $256 \times 256$ pixels.
It can be seen from Figs. \ref{Figure 4} and \ref{Figure 7} that LaMa \cite{9707077} is not good at repairing the large-area mask of high-resolution face.
Some low-quality images are generated with the high-resolution images and large-scale masks.
Since we use $256 \times 256$ images to train the LaMa \cite{9707077}, the image with $256 \times 256$ pixels can be repaired very well.
Based on this finding, we design a two-stage model. In the first stage, we use the original LaMa \cite{9707077} to generate the coarse image in $256 \times 256$ resolution.
In the second stage, we use bilinear to upsample the image and then use another LaMa-like model to refine the upsampled image.
We can see from Fig. \ref{Figure 4} that the two-stage model called GLaMa-2stage can generate more realistic faces.

\subsection{NTIRE 2022 Image Inpainting Challenge}
\label{sec:NTIRE 2022 Image Inpainting Challenge}
This work is proposed to participate in the NTIRE 2022 Image Inpainting Challenge Track 1 Unsupervised \cite{romero2022ntire}.
The objective of this challenge is to obtain a mask agnostic network solution capable of producing high-quality results for the best perceptual quality with respect to the ground truth.
Our results for the NTIRE 2022 challenge \cite{romero2022ntire} are denoted as AIIA (our team's name) to distinguish from the other teams in Tab. \ref{Table 5}.
It can be seen that our method goes significantly ahead of FID, LPIPS, PSNR and SSIM for Places2 \cite{zhou2017places}, ImageNet \cite{5206848} and WikiArt \cite{Saleh2015LargescaleCO} datasets.

\section{Conclusion}
\label{sec:Conclusion}
This study explores the mask generation strategy in image inpainting task.
Moreover, we introduce a joint spatial and frequency loss to reduce checkerboard effect and ripples.
The proposed GLaMa achieves significant performance improvements over LaMa \cite{9707077} without changing the model architecture. At the same time, the proposed method is more general and other methods can simply adopt the strategies we have explored.
Extensive ablation studies are performed to validate the effectiveness of our method, and comparative results on four public datasets demonstrates the state-of-the-art performance of our method.

\paragraph{Limitations}. There are also some failed examples when the input image are $512\times512$ pixels, as shown in Fig. \ref{Figure 7}. The results generated by GLaMa-2stage are sometimes blurred and unreasonable. LaMa \cite{9707077} and GLaMa also generate low-quality and unpleasant results.
In the future, we will incorporate the StyleGAN prior \cite{karras2019style} to assist the image inpainting task.

\noindent\textbf{Acknowledgments:} The research was supported by the National Natural Science Foundation of China (61971165), in part by the Fundamental Research Funds for the Central Universities (FRFCU5710050119), the Natural Science Foundation of Heilongjiang Province (YQ2020F004).

\newpage
{\small
\bibliographystyle{ieee_fullname}
\bibliography{egbib}
}

\end{document}